\documentclass[runningheads]{llncs}
\usepackage[T1]{fontenc}
\usepackage{graphicx}
\usepackage{amsmath,amssymb,amsfonts}
\usepackage{algorithmic}
\usepackage{graphicx}
\usepackage{textcomp}
\usepackage{multirow}
\usepackage{hhline}
\usepackage{xcolor}
\usepackage{url}
\urldef\slamasrurl\url{https://github.com/X-LANCE/SLAM-LLM/tree/main/examples/asr_librispeech}
\urldef\salmonnurl\url{https://huggingface.co/tsinghua-ee/SALMONN-7B}
\begin{document}
\title{Optimal Transport Regularization for Speech Text Alignment in Spoken Language Models
\thanks{
Work done when Wenze Xu interned at Mashang Consumer Finance Co., Ltd. \\
Wenze Xu and Chun Wang contributed equally to this work. \\
Corresponding author: Chun Wang, lukewang25@live.cn
}
}
\titlerunning{Optimal Transport Regularization for Speech Text Alignment in SLMs}
%
\author{
Wenze Xu\inst{1,2}\orcidID{0009-0005-2995-6270} \and
Chun Wang\inst{1}\orcidID{0009-0007-2290-910X} \and
Jiazhen Yu\inst{3}\orcidID{0009-0003-1282-3776} \and
Sheng Chen\inst{1}\orcidID{0009-0002-0397-3935} \and
Liang Gao\inst{1}\orcidID{0009-0005-6408-5271} \and
Weihong Deng\inst{1}\orcidID{0000-0001-5952-6996} 
}
\authorrunning{W. Xu et al.}
%
\institute{
Mashang Consumer Finance Co., Ltd., Chongqing, China \\
\email{lukewang25@live.cn} \\
\email{\{sheng.chen02, liang.gao01, weihong.deng\}@msxf.com} \and
The University of Sydney, Sydney, Australia \\
\email{wexu0327@uni.sydney.edu.au} \and
Macau University of Science and Technology, Macau SAR, China \\
\email{1210030170@student.must.edu.mo}
}
\maketitle              
\begin{abstract}
Spoken Language Models (SLMs), which extend Large Language Models (LLMs) to perceive speech inputs, have gained increasing attention for their potential to advance speech understanding tasks. However, despite recent progress, studies show that SLMs often struggle to generalize across datasets, even for trained languages and tasks, raising concerns about whether they process speech in a text-like manner as intended. A key challenge underlying this limitation is the modality gap between speech and text representations. The high variability in speech embeddings may allow SLMs to achieve strong in-domain performance by exploiting unintended speech variations, ultimately hindering generalization. To mitigate this modality gap, we introduce Optimal Transport Regularization (OTReg), a method that formulates speech-text alignment as an optimal transport problem and derives a regularization loss to improve SLM training. In each training iteration, OTReg first establishes a structured correspondence between speech and transcript embeddings by determining the optimal transport plan, then incorporates the regularization loss based on this transport plan to optimize SLMs in generating speech embeddings that align more effectively with transcript embeddings. OTReg is lightweight, requiring no additional labels or learnable parameters, and integrates seamlessly into existing SLM training procedures. Extensive multilingual ASR experiments demonstrate that OTReg enhances speech-text alignment, mitigates the modality gap, and consequently improves SLM generalization across diverse datasets.

\end{abstract}

\section{Introduction}
\label{sec:intro}
Large Language Models (LLMs)~\cite{Llama3,qwen25} have significantly advanced text-based language understanding and generation. Their success has driven progress in speech understanding research, leading to efforts to extend LLMs into Spoken Language Models (SLMs)~\cite{Qwen2Audio,tang2024salmonn}, which can directly process speech input—an essential and natural aspect of language representation.

One practical approach to building SLMs involves integrating a pretrained speech encoder with an LLM through an adapter module~\cite{arora2025landscapespokenlanguagemodels}. The adapter functions as a bridge, transforming speech embeddings into a representation compatible with the LLM’s input space, thereby enabling direct speech input processing. This method leverages the strengths of existing components, allowing the SLM to process speech inputs while preserving the LLM's advanced language understanding capabilities. Previous studies have shown that SLMs employing this approach achieve competitive performance in major speech understanding tasks, such as automatic speech recognition (ASR)~\cite{SLAMASR} and speech-to-text translation~\cite{10389705}, validating the effectiveness of adapter-based integration in SLM development.

Despite advancements in in-domain datasets, recent studies~\cite{fan2024alignformer,11010998} indicate that SLMs experience significant performance degradation across datasets, even when the spoken languages and tasks were included in training. These findings raise concerns that SLMs may not truly comprehend speech in a text-like manner as intended. Instead, their strong performance on in-domain data may partially result from exploiting unintended speech variations, suggesting overfitting~\cite{11010998}. Addressing these limitations is crucial for enhancing generalization and unlocking the full potential of SLMs in diverse speech applications.

\begin{figure}[htb]
  \centering
  \begin{minipage}[b]{1.0\linewidth}
    \centering
    \centerline{\includegraphics[width=0.5\linewidth]{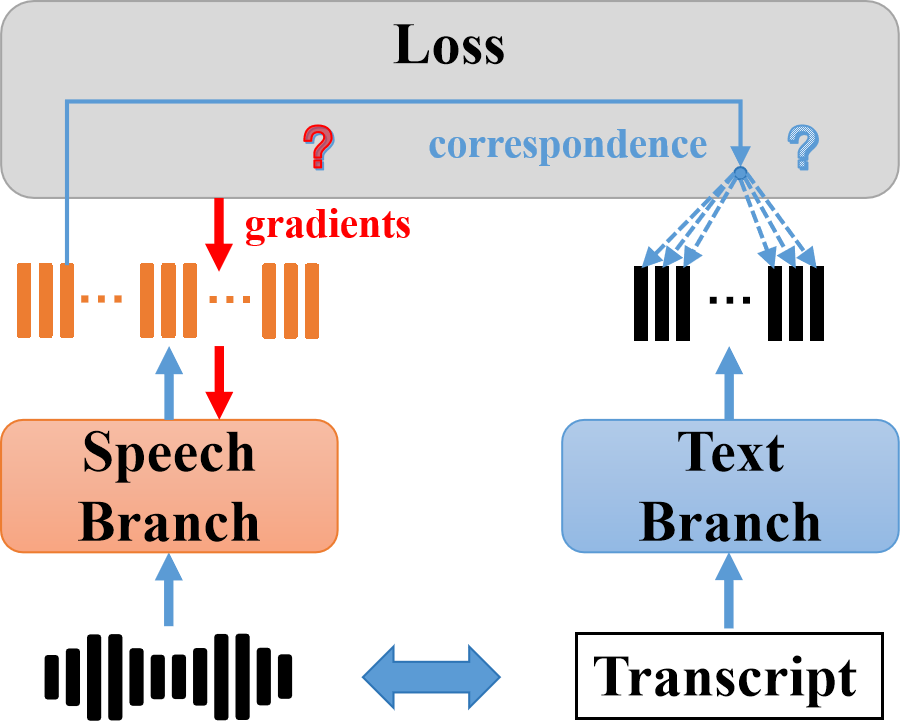}}
  \end{minipage}
  \caption{Challenges in speech-text alignment.}
  \label{fig:challenges}
\end{figure}

Recent studies~\cite{fan2024alignformer,pmlr-v202-le23a,tsiamas-etal-2024-pushing} suggest that the modality gap between speech and text is a key factor in this challenge. Because of high frame rates, speech embeddings are often substantially longer than their corresponding transcript embeddings. Furthermore, unlike transcript embeddings, which primarily encode linguistic content, speech embeddings also incorporate paralinguistic features such as pauses and variations in speech rate, adding complexity to their representation. These differences make speech more dynamic than text, increasing the likelihood that SLMs will capture irrelevant variations in speech rather than focusing solely on linguistic content. Addressing these issues requires precise speech-text alignment to reduce the modality gap, thereby enhancing SLM generalization.

Achieving precise speech-text alignment during SLM training presents two key challenges. As illustrated in Fig.~\ref{fig:challenges}, the first challenge is that while pairing speech with its transcript is straightforward, defining and establishing meaningful pointwise correspondences between heterogeneous speech and transcript embeddings remains difficult. The second challenge is designing a differentiable and robust loss function that optimizes the speech model to generate speech embeddings aligned with transcript embeddings based on the established pointwise correspondences. This is particularly challenging because dynamically established correspondences are often noisy and tend to change during training.

In this paper, to tackle these challenges, we introduce Optimal Transport Regularization (OTReg) as a novel approach to bridging the modality gap in SLMs. We formulate speech-text alignment as an optimal transport (OT) problem~\cite{torres2021surveyoptimaltransportmachine}, treating speech embeddings as the source and transcript embeddings as the target. In this framework, pointwise correspondences between the source and target are represented by a transport plan. Given a predefined cost matrix, OT determines the optimal transport plan by solving an optimization problem that minimizes the total transport cost. To encourage correspondences between similar embeddings, we define the transport cost based on cross-modal embedding similarity, ensuring that lower transport costs occur when embeddings exhibit higher similarity. Furthermore, to train SLMs to generate speech embeddings that better align with transcript embeddings, we derive a regularization loss from the optimal transport plan and incorporate it into the SLM training process. Each training iteration consists of two steps: first, estimating the optimal transport plan based on speech embeddings generated using the current SLM parameters; second, computing a regularization loss derived from the estimated transport plan and integrating it with the primary loss, jointly optimizing the SLM through backpropagation.

The proposed OTReg directly employs transcript embeddings as alignment targets, requires no additional annotated labels, and introduces no extra learnable parameters, ensuring seamless integration into the SLM training framework. Experimental results on multilingual ASR tasks demonstrate that OTReg effectively enhances speech-text alignment, reduces the modality gap, and consequently improves the generalization of SLMs across diverse datasets.

The remainder of this paper is structured as follows: Sec.~\ref{sec:pre} covers preliminaries, then Sec.~\ref{sec:relatedwork} reviews related work. The proposed method is detailed in Sec.~\ref{sec:method}. Then, Sec.~\ref{sec:exp} and Sec.~\ref{sec:res} present the experimental setup and results, respectively. Finally, Sec.~\ref{sec:con} concludes the paper.

\section{Preliminaries}
\label{sec:pre}


In this section, we provide a concise recap of CTC and Optimal Transport to facilitate understanding of related work and this study.
\subsection{Connectionist Temporal Classification}
\label{ssec:ctc}
Connectionist Temporal Classification (CTC) loss~\cite{CTC} is widely used in speech processing tasks to align speech embeddings with textual labels when annotated pointwise correspondence is unavailable. Instead of relying on predetermined alignments, CTC formulates the alignment problem by predicting a probability distribution over all possible label sequences derived from speech embeddings. The training objective is to maximize the cumulative probability of all sequences that, when appropriately collapsed, match the ground truth label sequences. To handle non-informative segments such as silence or pauses in speech, CTC introduces a special ``blank'' label (denoted as ``-''), allowing the model to accommodate timing and duration variability.

CTC-based compression~\cite{gaido-etal-2021-ctc} extends CTC to reduce redundancy in speech embeddings while preserving essential linguistic information. It leverages CTC predictions to merge consecutive embeddings corresponding to the same label and discard non-informative embeddings associated with the blank label. This approach enhances speech-to-text alignment and boosts computational efficiency by reducing the number of processed embeddings.

Although generally effective, CTC may be suboptimal for SLMs. The classifier used in CTC has a large number of parameters, making it prone to overfitting itself. Additionally, CTC relies on textual labels as intermediate targets, providing only indirect supervision. As a result, the generated speech embeddings may not align well with the LLM’s transcript embeddings in the metric space, creating a gap between the supervision signal and the ultimate goal of speech-text alignment in SLMs.

\subsection{Optimal Transport}
\label{ssec:ot}
Optimal Transport (OT)~\cite{torres2021surveyoptimaltransportmachine} provides a mathematical framework for discovering meaningful relationships between probability distributions by optimally transferring mass while preserving structural properties. It achieves this by solving a constrained optimization problem that minimizes a predefined cost function. The transport plan, which is the mathematical solution to this problem, specifies how mass is redistributed from the source to the target while minimizing transport cost, establishing meaningful correspondences between distributions.

A significant challenge in applying OT is its computational complexity, particularly in high-dimensional settings. To address this, techniques such as entropic regularization—exemplified by the Sinkhorn algorithm~\cite{10.5555/2999792.2999868}—have been developed. These methods introduce a regularization term that smooths the optimization landscape, enabling efficient computations through iterative scaling while preserving solution accuracy.

\section{Related Work}
\label{sec:relatedwork}

\subsection{Spoken Language Models}
\label{ssec:slm}
Prior studies on SLMs generally follow two main paradigms. One approach focuses on developing unified speech-text foundation models that natively process spoken language by jointly learning from both audio and text modalities. However, this approach typically requires extensive multimodal data and substantial computational resources, which limit its feasibility~\cite{arora2025landscapespokenlanguagemodels,WavChat}.

An alternative and increasingly popular approach enhances text-based LLMs with speech understanding capabilities. In this framework, a dedicated speech encoder is integrated with an LLM via an adapter module, aligning speech embeddings with the model's textual representation space. Recent studies have adopted this strategy to develop SLMs for specific tasks—such as ASR~\cite{SLAMASR} and speech-to-text translation~\cite{10389705}—as well as for general-purpose applications, exemplified by Qwen2-Audio~\cite{Qwen2Audio} and SALMONN~\cite{tang2024salmonn}. These models have demonstrated competitive performance across multiple tasks. Our work builds upon this approach, further refining speech and text alignment to enhance the generalization of SLMs.

\subsection{Speech-Text Cross-Modality Alignment}
\label{ssec:stcma}

%
Precise speech-text alignment is essential for bridging the modality gap between spoken and written language. Prior studies have explored explicit alignment methods to bring speech representations closer to text representations. Chuang et al.~\cite{chuang-etal-2020-worse} apply pretrained word embeddings as intermediate supervision, aligning speech embeddings by minimizing cosine distance. WACO~\cite{ouyang-etal-2023-waco} employs contrastive learning to enhance representation similarity between corresponding speech and text words while pushing apart non-corresponding representations in the embedding space. However, these methods struggle to accommodate unstable and noisy pointwise cross-modality correspondences during training. In scenarios where precise pointwise annotations are unavailable, Connectionist Temporal Classification (CTC)~\cite{gaido-etal-2021-ctc,10389705} has emerged as the most widely used approach (see~\cite{arora2025landscapespokenlanguagemodels} for further details). However, it relies on textual labels as intermediate targets, which are not well aligned with the ultimate goal of speech-text alignment in embedding space. More recently, Optimal Transport (OT) has been applied to establish correspondences. To accommodate the temporal nature of speech, CMOT~\cite{zhou-etal-2023-cmot} integrates a window strategy into OT to enforce monotonicity and locality when aligning speech sequences to text sequences, while Le et al.~\cite{pmlr-v202-le23a} introduces positional encoding to ensure that correspondences between the two sequences remain monotonic. However, due to inherent dynamics in speech—such as variations in speech rate and silence segments—these position-based constraints are difficult to determine. Therefore, in this work, we choose to use unique transcript embeddings as alignment targets rather than pursuing monotonicity.

\section{Method}
\label{sec:method}
This section begins with an overview of the SLM architecture, followed by a detailed explanation of the proposed Optimal Transport Regularization (OTReg). Finally, we integrate OTReg into the existing SLM training framework and present a two-stage training approach.

\begin{figure}[tb]
\centering
\begin{minipage}[b]{1.0\linewidth}
  \centering
  \centerline{\includegraphics[width=1.0\linewidth]{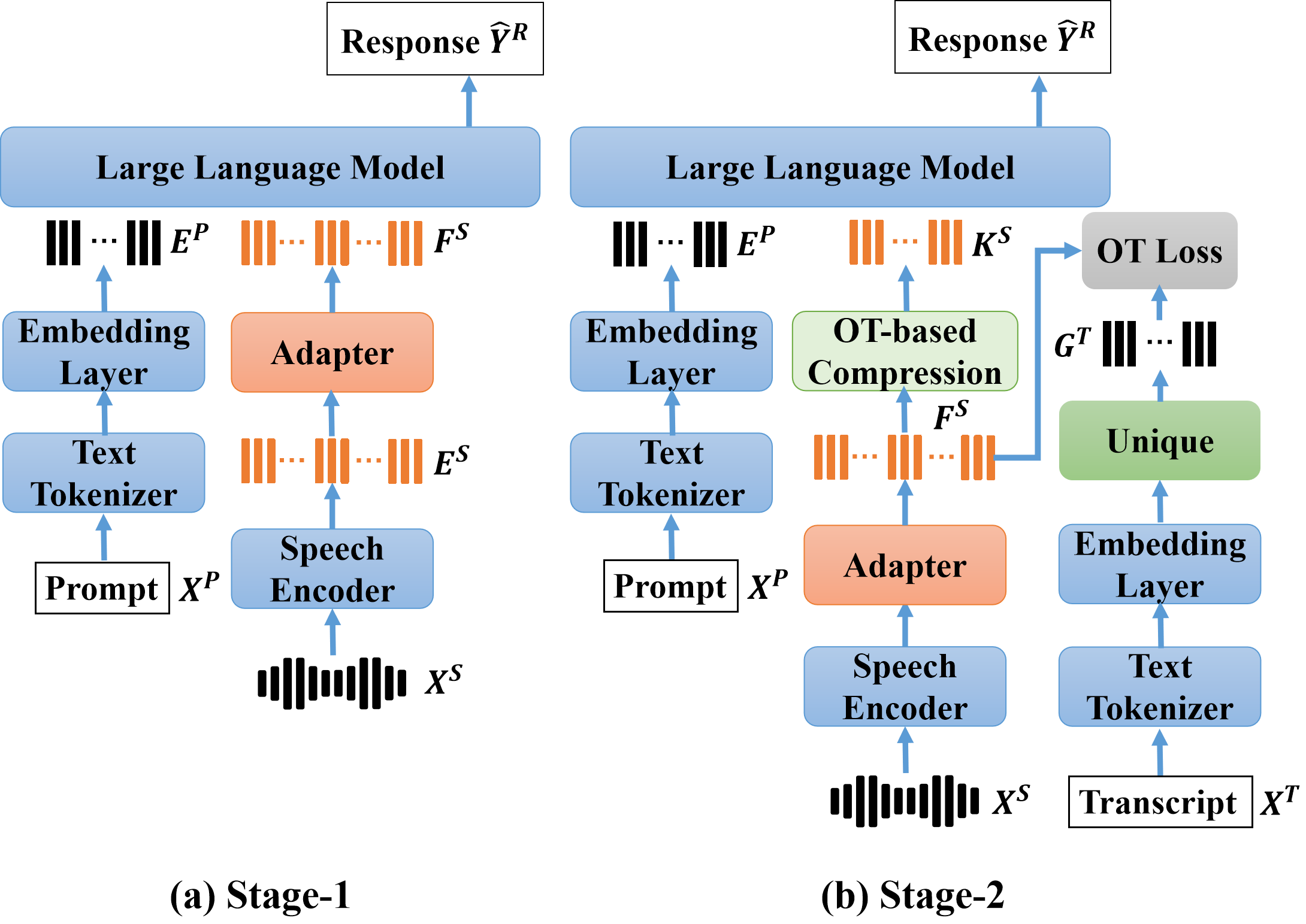}}
\end{minipage}
\caption{Overview of SLM training across Stage 1 and Stage 2. During training, blue modules remain frozen, orange modules are learnable, and green modules function without trainable parameters.}
\label{fig:stage12}
\end{figure}

%
\subsection{Model Architecture}
\label{ssec:modelarch}
As illustrated in Fig.~\ref{fig:stage12}(a), an SLM extends a text-based LLM by incorporating speech understanding capabilities and consists of three core components: a speech encoder, an adapter, and a text-based LLM, with its tokenizer and embedding layer exposed.

Unlike conventional LLMs that process only text, SLMs accept multimodal inputs, including a text prompt, $\mathbf{X}^{\mathbf{P}}$, and a speech signal, $\mathbf{X}^{\mathbf{S}}$. These inputs are encoded separately into distinct embeddings.

The text prompt, $\mathbf{X}^{\mathbf{P}}$, is processed using the LLM’s tokenizer and embedding layer to generate the prompt embeddings, $\mathbf{E}^{\mathbf{P}} \in \mathbb{R}^{n_p \times d_l}$: \begin{equation} \mathbf{E}^{\mathbf{P}} = \text{EmbedLayer}(\text{Tokenizer}(\mathbf{X}^{\mathbf{P}})), \end{equation} where $n_p$ denotes the sequence length, and $d_l$ represents the LLM’s embedding dimension.

Similarly, the speech signal, $\mathbf{X}^{\mathbf{S}}$, is processed through the speech encoder to produce the raw speech embeddings, $\mathbf{E}^{\mathbf{S}} \in \mathbb{R}^{n_s \times d_s}$: \begin{equation} \mathbf{E}^{\mathbf{S}} = \text{SpeechEncoder}(\mathbf{X}^{\mathbf{S}}), \end{equation} where $n_s$ and $d_s$ denote the sequence length and embedding dimension of the speech embeddings, respectively.

Next, the adapter transforms the raw speech embeddings, $\mathbf{E}^{\mathbf{S}}$, into transformed speech embeddings, $\mathbf{F}^{\mathbf{S}} \in \mathbb{R}^{n_a \times d_l}$, ensuring compatibility with the LLM’s embedding space. Here, $n_a$ represents the sequence length of the transformed speech embeddings.

Several adapter designs have been proposed in the literature~\cite{arora2025landscapespokenlanguagemodels}. Without loss of generality, we follow the approach presented in~\cite{SLAMASR} and implement the adapter using two linear layers with an intermediate ReLU activation function. The adapter has a hidden layer dimension, $d_h$, and is formulated as: \begin{equation} \label{eq:FS} \mathbf{F}^{\mathbf{S}} = \text{Linear}(\text{ReLU}(\text{Linear}(\mathbf{H}^{\mathbf{S}}))). \end{equation}

The intermediate representation, $\mathbf{H}^{\mathbf{S}} \in \mathbb{R}^{(n_s // k) \times (d_s \cdot k)}$, is a compact reformulation of $\mathbf{E}^{\mathbf{S}}$, where every $k$ consecutive embeddings, $\mathbf{e}^{\mathbf{S}}_{i}, \mathbf{e}^{\mathbf{S}}_{i+1}, \dots, \mathbf{e}^{\mathbf{S}}_{i+k-1}$, are concatenated into a single compact embedding, $\mathbf{h}^{\mathbf{S}}_{j}$.

Finally, the model integrates multimodal information and passes it into the LLM to generate a text response, $\mathbf{\hat{Y}}^{\mathbf{R}}$: \begin{equation} \label{eq:stage1} \text{Template}(\mathbf{E}^{\mathbf{P}}, \mathbf{F}^{\mathbf{S}}) \rightarrow \text{LLM} \rightarrow \mathbf{\hat{Y}}^{\mathbf{R}}, \end{equation} where Template refers to the instruction template used by the underlying LLM.

During supervised fine-tuning (SFT), the SLM is optimized by minimizing a cross-entropy (CE) loss function, $L_{CE}$, computed between the generated $\mathbf{\hat{Y}}^{\mathbf{R}}$ and the ground truth, $\mathbf{Y}^{\mathbf{R}}$. A common fine-tuning strategy involves updating only the adapter while keeping the speech encoder and LLM components frozen, thereby preserving their pre-trained capabilities.

\subsection{Optimal Transport Regularization}
\label{ssec:otreg}

\begin{figure}[tb]
\centering
\begin{minipage}[b]{1.0\linewidth}
  \centering
  \centerline{\includegraphics[width=0.8\linewidth]{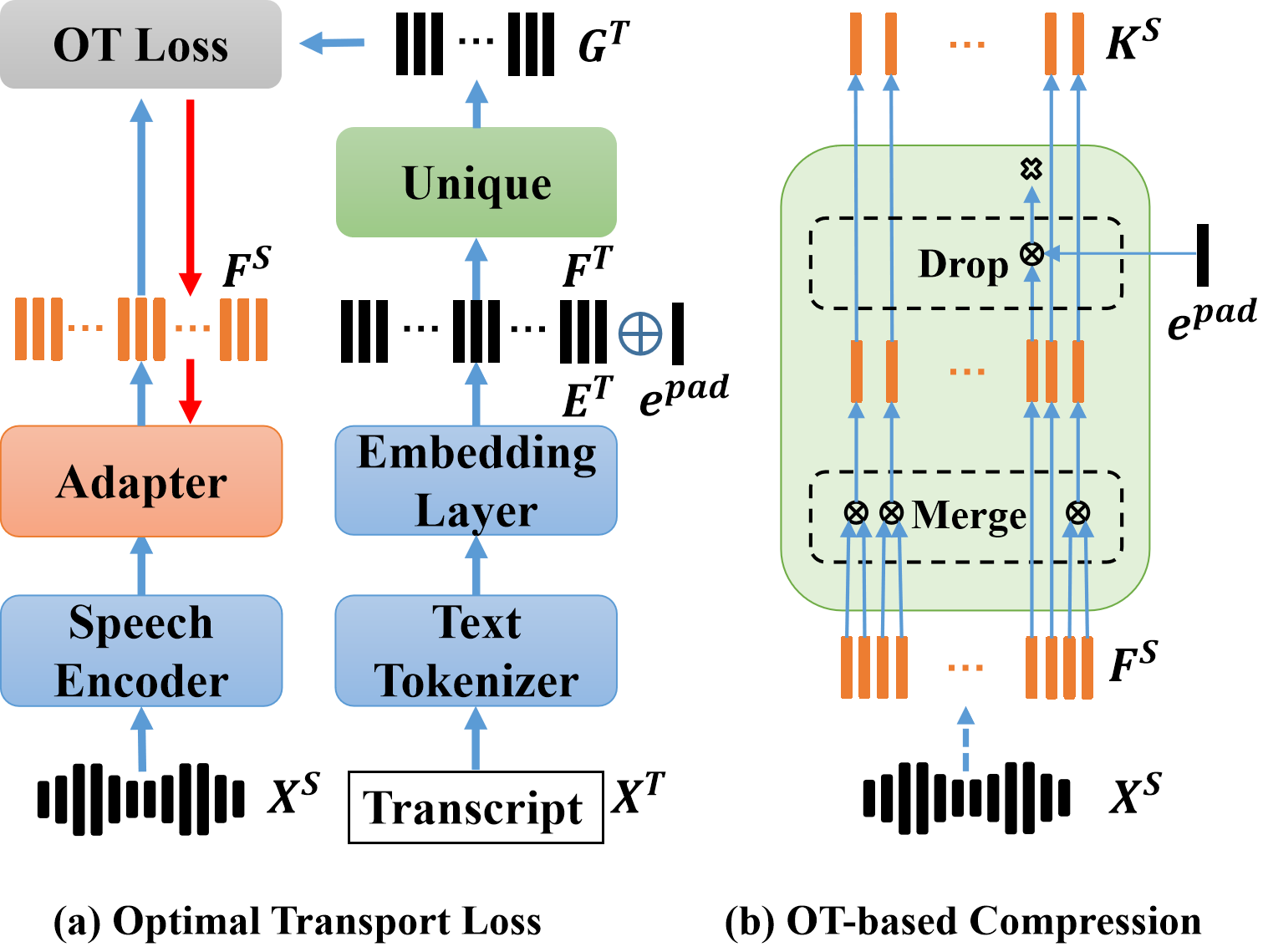}}
\end{minipage}
\caption{Illustration of OT loss and OT-based compression.}
\label{fig:ot}
\end{figure}

In the absence of annotated pointwise correspondences, we formulate the speech-text alignment problem as an OT problem. In our formulation, as illustrated in Fig.~\ref{fig:ot}(a), the transformed speech embeddings, $\mathbf{F}^{\mathbf{S}}$, serve as the source. For the target, transcripts often contain repeated patterns—such as "banana"—which can introduce ambiguity into the transport process. To address this issue, we derive unique transcript embeddings, $\mathbf{G}^{\mathbf{T}}$, from the speech transcript, $\mathbf{X}^{\mathbf{T}}$, using a dedicated processing procedure, as follows.

First, the transcript, $\mathbf{X}^{\mathbf{T}}$, is processed using the LLM's tokenizer and embedding layer to produce transcript embeddings, $\mathbf{E}^{\mathbf{T}} \in \mathbb{R}^{n_t \times d_l}$: \begin{equation} \mathbf{E}^{\mathbf{T}} = \text{EmbedLayer}(\text{Tokenizer}(\mathbf{X}^{\mathbf{T}})), \end{equation} where $n_t$ denotes the sequence length.

Second, to accommodate blank embeddings, $\mathbf{E}^{\mathbf{T}}$ is concatenated with the LLM's pad token embedding, $\mathbf{e}^{\text{pad}}$, to obtain the augmented transcript embeddings, $\mathbf{F}^{\mathbf{T}}$: \begin{equation} \mathbf{F}^{\mathbf{T}} = \text{Concat}([\mathbf{E}^{\mathbf{T}}; \mathbf{e}^{\text{pad}}]). \end{equation}

Third, we generate the set of unique embeddings, $\mathcal{S}^{\mathbf{T}}$, by extracting unique embeddings from $\mathbf{F}^{\mathbf{T}}$. Uniqueness is determined by comparing cosine similarities to filter out semantically equivalent embeddings. Since speech embeddings should align only with those present in the transcript, $\mathcal{S}^{\mathbf{T}}$—which encompasses all distinct transcript embeddings—serves as a robust target for the OT-based alignment process. To enhance usability, we concatenate the embeddings in $\mathcal{S}^{\mathbf{T}}$ into the unique transcript embeddings, $\mathbf{G}^{\mathbf{T}} \in \mathbb{R}^{n_g \times d_l}$, where $n_g$ represents the number of unique embeddings. Note that the order of unique embeddings in $\mathbf{G}^{\mathbf{T}}$ does not affect alignment, as OT is invariant to temporal order.

Another key component of the OT framework is the cost function, which quantifies the transport cost between embeddings from the source and target sequences. For each source embedding, $\mathbf{f}_i \in \mathbf{F}^{\mathbf{S}}$, and each target embedding, $\mathbf{g}_j \in \mathbf{G}^{\mathbf{T}}$, the cost is defined as: \begin{equation} \mathbf{C}_{ij} = 1 - \operatorname{CosineSimilarity}\left(\frac{\mathbf{f}_i}{\|\mathbf{f}_i\|_2}, \frac{\mathbf{g}_j}{\|\mathbf{g}_j\|2}\right), \end{equation} where $\operatorname{CosineSimilarity}$ returns values in the range $[-1, 1]$, ensuring similarity is computed using unit vectors. Consequently, $\mathbf{C}_{ij}$ falls within the range $[0, 2]$, with lower cost values occurring when transport happens between source and target embeddings of higher similarity, aligning with the objective of effective cross-modal alignment.

In a standard OT problem, the primary objective is to determine a transport plan, $\gamma \in \mathbb{R}^{n_a \times n_g}$, that minimizes the total transport cost while satisfying the given marginal constraints. In our approach, we further need to derive a regularization loss from the obtained transport plan to optimize the SLM, making differentiability essential. To achieve this, we employ entropic-regularized OT and solve it using the Sinkhorn algorithm~\cite{10.5555/2999792.2999868}.

Formally, the entropic-regularized OT problem is defined as: \begin{equation} \label{eq:ot_combined} \begin{aligned} \min_{\mathbf{\gamma}} \quad & \sum_{i=0}^{n_a-1} \sum_{j=0}^{n_g-1} \mathbf{\gamma}_{ij} \mathbf{C}_{ij} - \epsilon H(\mathbf{\gamma}) \\ \text{subject to} \quad & \mathbf{\gamma}_{ij} \ge 0, \quad \forall\, 0 \le i < n_a,\; 0 \le j < n_g, \\ & \sum_{j=0}^{n_g-1} \mathbf{\gamma}_{ij} = 1/n_a, \quad \forall\, 0 \le i < n_a, \\ & \sum_{i=0}^{n_a-1} \mathbf{\gamma}_{ij} = 1/n_g, \quad \forall\, 0 \le j < n_g. \end{aligned} \end{equation}
where $\epsilon > 0$ is a weight controlling the entropy term, and the Shannon entropy is defined as: \begin{equation} \label{eq:entropy} H(\mathbf{\gamma}) = -\sum_{i=0}^{n_a-1} \sum_{j=0}^{n_g-1} \mathbf{\gamma}_{ij} \log \mathbf{\gamma}_{ij}. \end{equation}

The optimal transport plan, $\hat{\mathbf{\gamma}}$, which minimizes \eqref{eq:ot_combined}, can be efficiently computed using the Sinkhorn algorithm.

Given the optimal transport plan, $\hat{\mathbf{\gamma}}$, we further derive a regularization loss consisting of two terms. The first term, the total transport cost, $L_{\text{cost}}$, quantifies the overall alignment quality via the total transport cost and is computed as \begin{equation} L_{\text{cost}}(\hat{\mathbf{\gamma}}) = \sum_{i=0}^{n_a-1} \sum_{j=0}^{n_g-1} \hat{\mathbf{\gamma}}_{ij} \mathbf{C}_{ij}. \end{equation}

For the second term, we introduce a sparsity constraint, \(L_{\text{spr}}\), to encourage a focused point-to-point transport plan, ensuring that each source speech embedding aligns primarily with a single transcript embedding. We compute this task-specific constraint using the L2 norm of each row in the row-normalized optimal transport plan, \(\hat{\mathbf{\gamma}}^{\text{row}}\), defined as
\begin{equation}
  L_{\text{spr}}(\hat{\mathbf{\gamma}}) = \frac{1}{n_a}\sum_{i=0}^{n_a-1} \left(1 - L_2\bigl(\hat{\mathbf{\gamma}}^{\text{row}}_{i:}\bigr)\right).
\end{equation}

Each row, $\hat{\boldsymbol{\gamma}}^{\text{row}}_{i:} = \frac{\hat{\boldsymbol{\gamma}}_{i:}}{\sum_{j=0}^{n_g - 1} \hat{\boldsymbol{\gamma}}_{ij}}$, forms a valid probability distribution over transcript embeddings. The term, $1 - L_2\bigl(\hat{\boldsymbol{\gamma}}^{\text{row}}_{i:}\bigr) \in [0, 1]$, is minimized when the probability distribution is highly concentrated (i.e., nearly one-hot), and the correspondence is nearly one-to-one. Consequently, minimizing this sparsity cost promotes a clear and distinct alignment between speech and transcript embeddings.

Finally, given the optimal transport plan, $\hat{\mathbf{\gamma}}$, the derived regularization loss, $L_{\text{OT}}$, is defined as \begin{equation} \label{eq:OT_loss} L_{\text{OT}}(\hat{\mathbf{\gamma}}) = L_{\text{cost}}(\hat{\mathbf{\gamma}}) + \lambda_{\text{spr}} L_{\text{spr}}(\hat{\mathbf{\gamma}}^{row}), \end{equation} where $\lambda_{\text{spr}} > 0$ is the weighting coefficient.

\subsection{OT-based Compression}
\label{ssec:otbcompreesion}
To reduce redundancy in the speech embeddings, $\mathbf{F}^{\mathbf{S}}$, the proposed OTReg introduces a content-aware compression method. The OT-based compression merges consecutive repetitive embeddings and removes non-informative embeddings while preserving essential content. Unlike CTC-based compression, which relies on classifier predictions to identify consecutive repeats and blanks, OT-based compression operates in a similarity-based manner, leveraging the fact that embeddings are trained using cosine similarity.

Specifically, as illustrated in Fig.~\ref{fig:ot}(b), the OT-based compression consists of two steps. In the merge step, speech embeddings are grouped into adjacent pairs (e.g., 0 and 1, 2 and 3, etc.) and merged when their similarity exceeds a predefined threshold, as they are considered semantically identical. Otherwise, the embeddings remain unchanged. Notably, merging is performed only on adjacent pairs to prevent excessive compression, which could distort sequences; for example, converting ``hheelllloo'' into ``helo'' is excessive. Furthermore, in the drop step, any speech embedding highly similar to the LLM's pad token embedding, $\mathbf{e}^{\text{pad}}$, is regarded as non-informative and removed. This OT-based compression procedure is fully differentiable and introduces no additional learnable parameters. 

The resulting condensed speech embeddings \begin{equation} \label{eq:KS} \mathbf{K}^{\mathbf{S}} = \text{OTCompression}(\mathbf{F}^{\mathbf{S}}) \end{equation} can then be combined with the prompt embeddings and passed into the LLM to generate a text response, as follows: \begin{equation} \label{eq:stage2} \text{Template}(\mathbf{E}^{\mathbf{P}}, \mathbf{K}^{\mathbf{S}}) \rightarrow \text{LLM} \rightarrow \mathbf{\hat{Y}}^{\mathbf{R}}. \end{equation}

\subsection{Two-Stage Training of the SLM}
\label{ssec:twostagetraining}
In this section, we integrate OTReg into the SLM training procedure and train the SLM using a two-stage process. As illustrated in Fig.~\ref{fig:stage12}(a), the first stage follows the standard supervised fine-tuning (SFT) approach. The SLM generates responses using~\eqref{eq:stage1} and is optimized for the next-token prediction task with the cross-entropy loss, $L_{\text{CE}}$, while only the adapter module is trainable. Notably, OTReg, including OT-based compression, is not applied during the first stage, as its effectiveness depends on similarity measures that require a well-initialized model. Upon completing this stage, the adapter can produce transformed speech embeddings that are approximately compatible with the LLM’s embedding space.

In stage two, as shown in Fig.~\ref{fig:stage12}(b), fine-tuning continues with the integration of the proposed OTReg, including OT-based compression. During each fine-tuning iteration, in addition to the SFT steps, the model also determines the transport plan and calculates the regularization loss, as follows:

\begin{enumerate} 
  \item Obtain transformed speech embeddings ($\mathbf{F}^{\mathbf{S}}$) using~\eqref{eq:FS}, based on the current SLM parameters. 
  \item Generate condensed speech embeddings ($\mathbf{K}^{\mathbf{S}}$) using~\eqref{eq:KS}. 
  \item Generate the model output ($\mathbf{\hat{Y}}^{\mathbf{R}}$) using~\eqref{eq:stage2}. 
  \item Evaluate performance by computing the cross-entropy loss ($L_{\text{CE}}$). 
  \item Compute the optimal transport plan ($\hat{\mathbf{\gamma}}$) by solving \eqref{eq:ot_combined}. 
  \item Calculate the regularization loss ($L_{\text{OT}}(\hat{\mathbf{\gamma}})$) according to~\eqref{eq:OT_loss}. 
  \item Update the SLM parameters via backpropagation using the total loss: \begin{equation} L_{\text{total}} = L_{\text{CE}} + \lambda_{OT} L_{\text{OT}}, \end{equation} where $\lambda_{OT} > 0$ is the weighting coefficient. \end{enumerate}

This two-stage approach improves speech-text alignment, reduces the modality gap, and thereby enhances generalization.

Compared to CTC, which provides a framework suitable for broader scenarios, our proposed OTReg is specifically designed for SLMs. OTReg establishes direct and dense supervision by leveraging unique transcript embeddings as concrete alignment targets, enabling similarity-based cross-modal comparison. In contrast, CTC relies on textual labels as intermediate targets, providing only indirect supervision. Consequently, the generated speech embeddings may not align well with the transcript embeddings of the LLM in the metric space, leaving a gap between the supervision signal and the ultimate goal of speech-text alignment in SLMs. 

\section{Experiments}
\label{sec:exp}

\subsection{Datasets}
\label{ssec:datasets}
To evaluate the effectiveness of our proposed OTReg on SLMs, we consider the multilingual ASR task—a primary challenge in speech understanding—and assess our method using multiple benchmark datasets. For in-domain training and evaluation, we utilize the English data of the LibriSpeech (LR960) dataset~\cite{7178964} along with the French and Spanish subsets of the Multilingual LibriSpeech (MLS) dataset~\cite{pratap20_interspeech}. For cross-domain evaluation, we leverage the English, French, and Spanish datasets from CoVoST-2~\cite{wang2020covost2massivelymultilingual} as well as the FLEURS dataset~\cite{10023141}. All datasets are used with their official splits. 

\subsection{Model Implementation}
\label{ssec:modelimpl}
In this work, we utilize Whisper-large-v3~\cite{whisper} as the speech encoder and Qwen2.5-7B-Instruction~\cite{qwen25} as the LLM. Following~\cite{SLAMASR}, the adapter is designed with a hidden layer dimension of $d_h=2048$ and a downsampling rate of $k=5$, resulting in transformed speech embeddings, $\mathbf{F}^{\mathbf{S}}$, at a frequency of 10 Hz. For compression, we apply a cosine similarity threshold of 0.9. The prompt, $\mathbf{X}^{\mathbf{P}}$, is set to ``Write down the user's content word for word in \{language\}, without incorporating any other details.'', where \{language\} acts as a placeholder whose value is determined by the input sample language. Throughout the paper, we set $\lambda_{\text{spr}}=0.1$ for the loss function while ablating the value of $\lambda_{OT}$.

\subsection{Training Details}
All experiments were conducted using four A800-80GB GPUs with a batch size of 48. Optimization was performed using the AdamW optimizer with a peak learning rate of 1e-4. A cosine decay learning rate scheduler was applied, gradually reducing the learning rate to a minimum of 1e-6. For all experiments, Stage-1 training was run for 2 epochs, followed by Stage-2 training for an additional 3 epochs. Model checkpoints were selected based on the lowest validation loss.

\section{Results}
\label{sec:res}
In this section, we first evaluate our method on the English ASR task, comparing it with recent SLMs. Next, we assess its performance on the multilingual ASR task to demonstrate its effectiveness across different languages. For both tasks, we use the Whisper normalizer~\cite{whisper} to preprocess transcripts and evaluate performance using the standard Word Error Rate (WER) metric.

\begin{table}[htb]
  \centering
  \caption{English ASR results.}
\begin{tabular}{|l|cccc|}
\hline
\multicolumn{1}{|c|}{\multirow{3}{*}{\textbf{Model}}} &
  \multicolumn{2}{c|}{\textbf{In domain (WER$\downarrow$)}} &
  \multicolumn{2}{c|}{\textbf{Cross domain (WER$\downarrow$)}} \\ \cline{2-5} 
\multicolumn{1}{|c|}{} &
  \multicolumn{1}{c|}{\multirow{2}{*}{\begin{tabular}[c]{@{}c@{}}LS960\\ (clean)\end{tabular}}} &
  \multicolumn{1}{c|}{\multirow{2}{*}{\begin{tabular}[c]{@{}c@{}}LS960\\ (other)\end{tabular}}} &
  \multicolumn{1}{c|}{\multirow{2}{*}{\begin{tabular}[c]{@{}c@{}}CoVoST-2\\ (test)\end{tabular}}} &
  \multirow{2}{*}{\begin{tabular}[c]{@{}c@{}}FLEURS\\ (test)\end{tabular}} \\
\multicolumn{1}{|c|}{} &
  \multicolumn{1}{c|}{} &
  \multicolumn{1}{c|}{} &
  \multicolumn{1}{c|}{} &
   \\ \hline
SLAM-ASR~\cite{SLAMASR} &
  \multicolumn{1}{c|}{2.43} &
  \multicolumn{1}{c|}{\textbf{4.91}} &
  \multicolumn{1}{c|}{28.66} &
  8.88 \\ \hline
SALMONN~\cite{tang2024salmonn} &
  \multicolumn{1}{c|}{4.20} &
  \multicolumn{1}{c|}{7.18} &
  \multicolumn{1}{c|}{\textbf{12.78}} &
  \textbf{6.20} \\ \hhline{|=|=|=|=|=|}
Base &
  \multicolumn{1}{c|}{\textbf{2.09}} &
  \multicolumn{1}{c|}{\textit{\textbf{4.96}}} &
  \multicolumn{1}{c|}{16.06} &
  7.63 \\ \hline
Base+CTC&
  \multicolumn{1}{c|}{2.44} &
  \multicolumn{1}{c|}{5.00} &
  \multicolumn{1}{c|}{18.16} &
  7.71 \\ \hline
Base+OTReg($\lambda_{OT}$=0.1) &
  \multicolumn{1}{c|}{2.28} &
  \multicolumn{1}{c|}{5.19} &
  \multicolumn{1}{c|}{15.59} &
  8.23 \\ \hline
Base+OTReg($\lambda_{OT}$=0.3) &
  \multicolumn{1}{c|}{\textit{\textbf{2.19}}} &
  \multicolumn{1}{c|}{5.24} &
  \multicolumn{1}{c|}{\textit{\textbf{13.12}}} &
  \textit{\textbf{6.90}} \\ \hline
Base+OTReg($\lambda_{OT}$=0.5) &
  \multicolumn{1}{c|}{2.20} &
  \multicolumn{1}{c|}{5.24} &
  \multicolumn{1}{c|}{13.67} &
  7.10 \\ \hline
Base+OTReg($\lambda_{OT}$=0.7) &
  \multicolumn{1}{c|}{2.28} &
  \multicolumn{1}{c|}{5.31} &
  \multicolumn{1}{c|}{13.38} &
  7.10 \\ \hline
Base+OTReg($\lambda_{OT}$=1.0) &
  \multicolumn{1}{c|}{2.21} &
  \multicolumn{1}{c|}{5.37} &
  \multicolumn{1}{c|}{13.17} &
  7.12 \\ \hline
\end{tabular}
\label{tab:ASR}
\end{table}

\begin{figure}[tb]
\centering
\begin{minipage}[b]{1.0\linewidth}
  \centering
  \centerline{\includegraphics[width=0.55\linewidth]{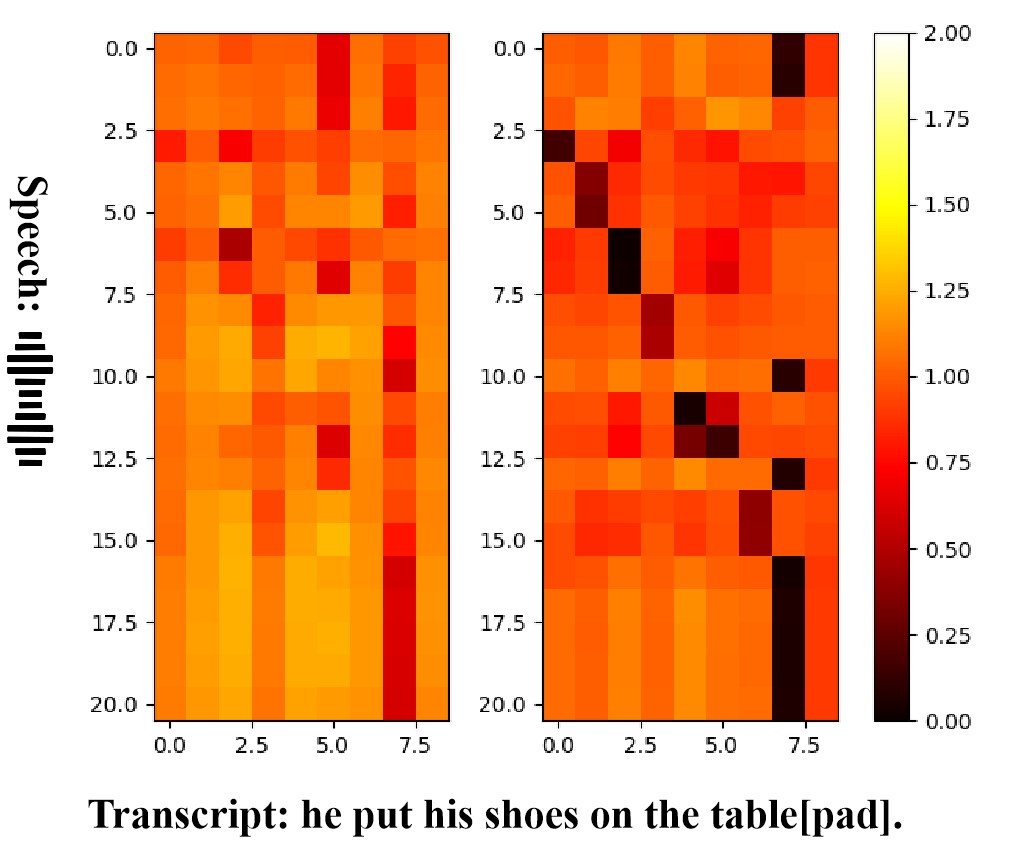}}
\end{minipage}
\caption{Distance between speech and transcript embeddings: lower values correspond to higher similarity.}
\label{fig:cost}
\end{figure}

\begin{table*}[tb]
  \centering
  \caption{Multilingual ASR results.}
\begin{tabular}{|l|cccc|cccccc|}
\hline
\multicolumn{1}{|c|}{\multirow{2}{*}{\textbf{Model}}} &
  \multicolumn{4}{c|}{\textbf{In Domain (WER$\downarrow$)}} &
  \multicolumn{6}{c|}{\textbf{Cross Domain (WER$\downarrow$)}} \\ \cline{2-11} 
\multicolumn{1}{|c|}{} &
  \multicolumn{2}{c|}{LS960} &
  \multicolumn{2}{c|}{MLS} &
  \multicolumn{3}{c|}{CoVoST-2} &
  \multicolumn{3}{c|}{FLEURS} \\ \hline
\multicolumn{1}{|c|}{} &
  \multicolumn{1}{c|}{En(clean)} &
  \multicolumn{1}{c|}{En(other)} &
  \multicolumn{1}{c|}{Es} &
  Fr &
  \multicolumn{1}{c|}{En} &
  \multicolumn{1}{c|}{Es} &
  \multicolumn{1}{c|}{Fr} &
  \multicolumn{1}{c|}{En} &
  \multicolumn{1}{c|}{Es} &
  Fr \\ \hline
Base &
  \multicolumn{1}{c|}{\textbf{2.17}} &
  \multicolumn{1}{c|}{\textbf{4.97}} &
  \multicolumn{1}{c|}{4.79} &
  \textbf{5.67} &
  \multicolumn{1}{c|}{16.21} &
  \multicolumn{1}{c|}{12.42} &
  \multicolumn{1}{c|}{19.33} &
  \multicolumn{1}{c|}{8.30} &
  \multicolumn{1}{c|}{7.78} &
  \textbf{10.76} \\ \hline
Base+CTC &
  \multicolumn{1}{c|}{2.18} &
  \multicolumn{1}{c|}{5.30} &
  \multicolumn{1}{c|}{\textbf{4.63}} &
  5.71 &
  \multicolumn{1}{c|}{15.45} &
  \multicolumn{1}{c|}{12.72} &
  \multicolumn{1}{c|}{18.57} &
  \multicolumn{1}{c|}{7.44} &
  \multicolumn{1}{c|}{\textbf{7.77}} &
  11.30 \\ \hline
Base+OTReg &
  \multicolumn{1}{c|}{2.35} &
  \multicolumn{1}{c|}{5.26} &
  \multicolumn{1}{c|}{5.03} &
  6.46 &
  \multicolumn{1}{c|}{\textbf{13.03}} &
  \multicolumn{1}{c|}{\textbf{9.54}} &
  \multicolumn{1}{c|}{\textbf{15.61}} &
  \multicolumn{1}{c|}{\textbf{6.92}} &
  \multicolumn{1}{c|}{7.86} &
  11.13 \\ \hline
\end{tabular}
\label{tab:multilingualASR}
\end{table*}

Tab.~\ref{tab:ASR} summarizes the results of the English ASR task, revealing several key findings. First, we evaluate two recent SLMs. SLAM-ASR\footnote{\slamasrurl; choose  wavlm\_large\_linear\_vicuna\_7b}~\cite{SLAMASR}, trained on LS960, suffers a significant performance drop across different datasets. In contrast, SALMONN\footnote{\salmonnurl}~\cite{tang2024salmonn}, a generic SLM trained on massive data and tasks, achieves considerably better results on CoVoST-2 and FLEURS but performs poorly on LS960. These results confirm the generalization issue of SLMs~\cite{11010998} and indicate that it cannot be eliminated simply by training with more data and tasks. 

Second, we develop our SLM, which retains the architectural foundation of SLAM-ASR but integrates more advanced components, including Whisper-large-v3 and Qwen2.5-7B-Instruction. We train multiple SLM variants using the proposed two-stage approach, applying either CTC or OTReg in the second stage, while Base is trained solely with CE loss. Results indicate that both CTC and OTReg achieve balanced performance across domains, performing comparably to SLAM-ASR on in-domain datasets while approaching SALMONN’s results in cross-domain evaluations. Furthermore, OTReg demonstrates robustness to loss weighting, yielding consistent results, with the best cross-domain performance observed at $\lambda_{OT} = 0.3$.

Following~\cite{audiochatllama,11010998}, we further illustrate the pairwise distance between speech embeddings, $\mathbf{F}^{\mathbf{S}}$, and their corresponding transcript embeddings, $\mathbf{E}^{\mathbf{T}}$. A test-set sample is shown in Fig.~\ref{fig:cost}, where Base-SLM (left) generates misaligned speech embeddings, suggesting that it may achieve good in-domain performance by leveraging speech variations. In contrast, with OTReg (right), the SLM produces speech embeddings that are well-aligned with transcript embeddings, mapping non-informative segments to the final pad embedding. This effectively reduces the modality gap, enabling the SLM to interpret speech in a text-like manner as intended while enhancing its ability to generalize across domains.

The results of the Multilingual ASR task are summarized in Tab.~\ref{tab:multilingualASR}. Similar to the English ASR cases, while Base-SLM has already demonstrated strong performance across datasets, the proposed OTReg ($\lambda_{OT} = 0.3$) further enhances SLM performance, particularly on cross-domain datasets for all three languages, reinforcing OTReg’s effectiveness in improving SLM's generalization across diverse linguistic scenarios.

\section{Conclusion}
\label{sec:con}
In this work, we address the challenge of cross-dataset generalization in SLMs by bridging the gap between speech and text modalities. We introduce Optimal Transport Regularization (OTReg), an efficient parameter-free method that seamlessly integrates into SLM training, refining speech-text alignment and enhancing SLMs’ generalization across datasets. Looking ahead, extending OTReg to broader speech understanding applications, such as zero-shot language settings, presents a promising direction for future research.

%
%
%
\bibliographystyle{splncs04}
\bibliography{submission}

\begin{thebibliography}{10}
\providecommand{\url}[1]{\texttt{#1}}
\providecommand{\urlprefix}{URL }
\providecommand{\doi}[1]{https://doi.org/#1}

\bibitem{arora2025landscapespokenlanguagemodels}
Arora, S., et~al.: On the landscape of spoken language models: A comprehensive
  survey. arXiv:2504.08528  (2025)

\bibitem{Qwen2Audio}
Chu, Y., et~al.: Qwen2-audio technical report. arXiv:2407.10759  (2024)

\bibitem{chuang-etal-2020-worse}
Chuang, S.P., Sung, T.W., Liu, A.H., Lee, H.y.: Worse {WER}, but better {BLEU}?
  leveraging word embedding as intermediate in multitask end-to-end speech
  translation. In: Proceedings of the 58th Annual Meeting of the Association
  for Computational Linguistics. pp. 5998--6003 (Jul 2020)

\bibitem{10023141}
Conneau, A., et~al.: Fleurs: Few-shot learning evaluation of universal
  representations of speech. In: 2022 IEEE Spoken Language Technology Workshop
  (SLT). pp. 798--805 (2023)

\bibitem{10.5555/2999792.2999868}
Cuturi, M.: Sinkhorn distances: lightspeed computation of optimal transport.
  In: Proceedings of the 27th International Conference on Neural Information
  Processing Systems - Volume 2. p. 2292–2300. NIPS'13 (2013)

\bibitem{Llama3}
Dubey, A., et~al.: The llama 3 herd of models. arXiv:2407.21783  (2024)

\bibitem{fan2024alignformer}
Fan, R., et~al.: Alignformer: Modality matching can achieve better zero-shot
  instruction-following speech-llm. arXiv:2412.01145  (2024)

\bibitem{audiochatllama}
Fathullah, Y., et~al.: {A}udio{C}hat{L}lama: Towards general-purpose speech
  abilities for {LLM}s. In: Proceedings of the 2024 Conference of the North
  American Chapter of the Association for Computational Linguistics: Human
  Language Technologies (Volume 1: Long Papers). pp. 5522--5532. Mexico City,
  Mexico (Jun 2024)

\bibitem{gaido-etal-2021-ctc}
Gaido, M., et~al.: {CTC}-based compression for direct speech translation. In:
  Proceedings of the 16th Conference of the European Chapter of the Association
  for Computational Linguistics: Main Volume. pp. 690--696 (Apr 2021)

\bibitem{CTC}
Graves, A., Fern\'{a}ndez, S., Gomez, F., Schmidhuber, J.: Connectionist
  temporal classification: labelling unsegmented sequence data with recurrent
  neural networks. In: Proceedings of the 23rd International Conference on
  Machine Learning. p. 369–376. ICML '06 (2006)

\bibitem{WavChat}
Ji, S., et~al.: Wavchat: A survey of spoken dialogue models. arXiv:2411.13577
  (2024)

\bibitem{11010998}
Kumar, S., et~al.: Performance evaluation of slam-asr: The good, the bad, the
  ugly, and the way forward. In: 2025 IEEE International Conference on
  Acoustics, Speech, and Signal Processing Workshops (ICASSPW). pp.~1--5 (2025)

\bibitem{pmlr-v202-le23a}
Le, P.H., Gong, H., Wang, C., Pino, J., Lecouteux, B., Schwab, D.: Pre-training
  for speech translation: {CTC} meets optimal transport. In: Proceedings of the
  40th International Conference on Machine Learning. Proceedings of Machine
  Learning Research, vol.~202, pp. 18667--18685 (23--29 Jul 2023)

\bibitem{SLAMASR}
Ma, Z., et~al.: Speech recognition meets large language model: Benchmarking,
  models, and exploration. Proceedings of the AAAI Conference on Artificial
  Intelligence (23),  24840--24848 (Apr 2025)

\bibitem{ouyang-etal-2023-waco}
Ouyang, S., Ye, R., Li, L.: {WACO}: Word-aligned contrastive learning for
  speech translation. In: Proceedings of the 61st Annual Meeting of the
  Association for Computational Linguistics (Volume 1: Long Papers). pp.
  3891--3907 (Jul 2023)

\bibitem{7178964}
Panayotov, V., Chen, G., Povey, D., Khudanpur, S.: Librispeech: An asr corpus
  based on public domain audio books. In: 2015 IEEE International Conference on
  Acoustics, Speech and Signal Processing (ICASSP). pp. 5206--5210 (2015)

\bibitem{pratap20_interspeech}
Pratap, V., Xu, Q., Sriram, A., Synnaeve, G., Collobert, R.: Mls: A large-scale
  multilingual dataset for speech research. In: Interspeech 2020. pp.
  2757--2761 (2020)

\bibitem{qwen25}
{Qwen Team}: Qwen2.5: A party of foundation models (September 2024)

\bibitem{whisper}
Radford, A., Kim, J.W., Xu, T., Brockman, G., Mcleavey, C., Sutskever, I.:
  Robust speech recognition via large-scale weak supervision. In: Proceedings
  of the 40th International Conference on Machine Learning. Proceedings of
  Machine Learning Research, vol.~202, pp. 28492--28518 (23--29 Jul 2023)

\bibitem{tang2024salmonn}
Tang, C., et~al.: {SALMONN}: Towards generic hearing abilities for large
  language models. In: The Twelfth International Conference on Learning
  Representations (2024)

\bibitem{torres2021surveyoptimaltransportmachine}
Torres, L.C., Pereira, L.M., Amini, M.H.: A survey on optimal transport for
  machine learning: Theory and applications. arXiv:2106.01963  (2021)

\bibitem{tsiamas-etal-2024-pushing}
Tsiamas, I., G{\'a}llego, G.I., Fonollosa, J.A.R., Costa-juss{\`a}, M.R.:
  Pushing the limits of zero-shot end-to-end speech translation. In: Findings
  of the Association for Computational Linguistics: ACL 2024. pp. 14245--14267
  (Aug 2024)

\bibitem{wang2020covost2massivelymultilingual}
Wang, C., Wu, A., Pino, J.: Covost 2 and massively multilingual speech-to-text
  translation. arXiv:2007.10310  (2020)

\bibitem{10389705}
Wu, J., et~al.: On decoder-only architecture for speech-to-text and large
  language model integration. In: 2023 IEEE Automatic Speech Recognition and
  Understanding Workshop (ASRU). pp.~1--8 (2023)

\bibitem{zhou-etal-2023-cmot}
Zhou, Y., Fang, Q., Feng, Y.: {CMOT}: Cross-modal mixup via optimal transport
  for speech translation. In: Proceedings of the 61st Annual Meeting of the
  Association for Computational Linguistics (Volume 1: Long Papers). pp.
  7873--7887 (Jul 2023)

\end{thebibliography}
\end{document}